# Training LDCRF model on unsegmented sequences using Connectionist Temporal Classification

[1]Amir Ahooye Atashin, [2]Kamaledin Ghiasi-Shirazi, [3]Ahad Harati
Department of Computer Engineering
Ferdowsi University of Mashhad
Mashhad, Iran
[1]amir.atashin@stu.um.ac.ir
[2]k.ghiasi@um.ac.ir
[3]a.harati@um.ac.ir

*Abstract.* **Many machine learning problems such as speech recognition, gesture recognition, and handwriting recognition are concerned with simultaneous segmentation and labeling of sequence data. Latent-dynamic conditional random field (LDCRF) is a well-known discriminative method that has been successfully used for this task. However, LDCRF can only be trained with pre-segmented data sequences in which the label of each frame is available apriori. In the realm of neural networks, the invention of connectionist temporal classification (CTC) made it possible to train recurrent neural networks on unsegmented sequences with great success. In this paper, we use CTC to train an LDCRF model on unsegmented sequences. Experimental results on two gesture recognition tasks show that the proposed method outperforms LDCRFs, hidden Markov models, and conditional random fields.**

## I. Introduction

Labeling and segmenting of data sequences is a common problem in sequence classification tasks [2]. It arises in many real-world applications such as gesture recognition, speech recognition, and handwriting recognition.

Probabilistic graphical models (PGMs) offer a powerful framework for machine learning in which the stochastic relations between several random variables is expressed economically by the use of graphical structures [3]. PGMs are divided into two categories: undirected graphical models or Markov random fields (MRFs) and directed graphical models or Bayesian networks (BNs). Conditional random fields (CRFs) [4], a variant of MRFs, and hidden Markov models (HMMs) [5], a variant of BNs, are two well-known graphical models which are widely used in sequence classification problems. HMM is a generative model which models the joint distribution of input sequences and their corresponding labels, but, CRF is a discriminative model which models the conditional distribution of label given the data sequence.

Latent-dynamic conditional random fields (LDCRFs) [6] are a type of CRFs which are used for structured prediction problems. It was proposed to be utilized in the particular case of sequence classification problems in which a label must be assigned to each frame of an unsegmented sequence. Learning procedure for these models is discriminative and carried out on a frame-wise labeled training set.

LDCRF model (Figure 1.b) can be described as a generalized form of chain-structured CRF model (Figure 1.a) in which hidden variables are incorporated into the CRF to learn various structures of class labels. Hidden conditional random fields (HCRFs) [1] are another variant of CRF which includes hidden variables. By incorporating hidden variables, HCRFs can capture intrinsic relations between sub-structures of class labels. In addition, to capturing intrinsic relations, the LDCRF also using those latent variables for learning extrinsic dynamics between class labels. Recent works have shown that LDCRF model and its variants are the proper methods for modeling data sequences. They are mostly applied to the sequence classification problems in the field of human-computer interaction (HCI) such as sign language spotting and human action recognition problems [7-13].

Although LDCRF can be employed for segmenting and tagging of unsegmented sequences, it requires a set of pre-segmented sequences during training. However, for many real-world data sets in sequence classification tasks, the labeled segmented sequence is not available, and so the LDCRF model cannot be directly trained on them.

Graves et all. [14] introduced a method called "connectionist temporal classification" (CTC) which can be used for training RNNs directly on unsegmented sequences. They used CTC as a layer on top of recurrent neural networks such as LSTM [15] and then applied it to the sequence classification problems such as online/offline unconstrained handwriting recognition. Their experiments showed that their method attained a significant improvement in classification performance over earlier methods [16-19].

In this paper, we use CTC to remove the need of pre-segmented sequences for training LDCRF and present a method for training LDCRFs directly on unsegmented sequences. The next section reviews some preliminary materials and introduces some definitions. Section III describes the motivations of the proposed method, its mathematical formulation, and its training algorithm. Section IV introduces the datasets used in our experiments for sequence classification. In Section V we experimentally compare the proposed CTC-LDCRF method with LDCRF, CRF, and HMM models. Section VI gives a direction for future works and concludes this paper.

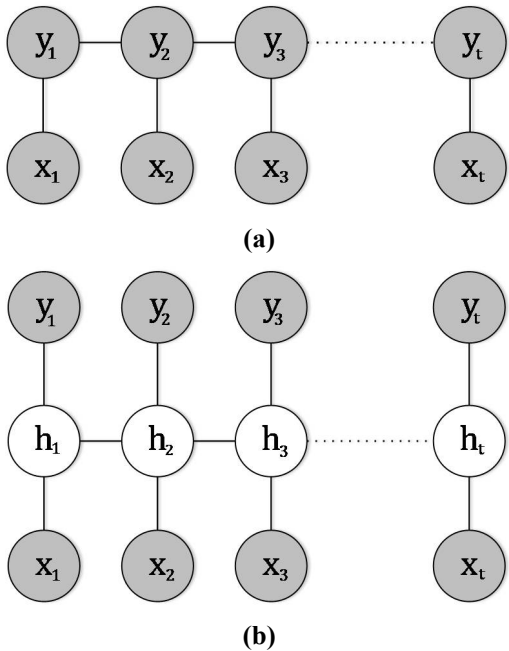

Figure 1. **(a)** Conditional random fields **(b)** Latent-dynamic conditional random fields. Gray nodes are indicative of observed variables.

## II. PRELIMINARIES

In this section, we first briefly explain the temporal classification task in general, and then review the LDCRF and CTC methods.

### A. Temporal Classification

In temporal classification, each input sequence is associated with a sequence of labels where the alignment between them is unknown [2, 20] . Let $S = \{(x_1, y_1), ..., (x_n, y_n)\}$ be a set of training samples where each sample $s_i = (x_i, y_i)$ consists of an input data sequence $x_i = \{x_{i1}, x_{i2}, ..., x_{it_i}\}$ with length $t_i$ and its corresponding label sequence $y_i = \{y_{i1}, y_{i2}, ..., y_{im_i}\}$ with length $m_i$ which is less than or equal to the $t_i$ ($m_i \leq t_i$). Each $x_{ij}$ is a d-dimensional feature vector and each $y_{ij}$ is a member of set $\mathcal{Y}$ which is the set consisting of all possible class labels. The objective of the temporal classification task is to learn the mapping between input data sequences and their corresponding label sequences.

### B. Latent-dynamic conditional random field (LDCRF)

LDCRF is an undirected graphical model that has been developed for the particular case of sequence classification problems where the goal is to predict a label at each frame for a given data sequence (Figure 1.b).

Let $S$ be a training set for the LDCRF that consists of $n$ samples. For each training sample in $S$, the class label sequence is as long as data sequence (i. e. $t_i = u_i \, for \, i = 1 ... n$).

In LDCRF, there is a sequence of hidden variables (i. e. $h_i = \{h_{i1}, h_{i2}, ..., h_{it_i}\}$) for each sample which not observed in $S$. Each $h_{ij}$ is a member of a set $\mathcal{H}_{y_j}$ of all possible hidden states for the class label $y_j$. In the LDCRF each class label is restricted to have a disjoint set of hidden states from other classes. Also set $\mathcal{H}$ defined as a set of all possible hidden states and it is a union of all $\mathcal{H}_y$ ($\mathcal{H} = \cup_y \mathcal{H}_y$).

Given the above definition, the LDCRF model with the parameter set $\theta$ is defined as latent conditional probability model, and for one particular training sample $(x, y)$ it can be written as:

$$P(y|x;\theta) = \sum_{h \in \mathcal{H}} P(y|h,x;\theta) P(h|x;\theta)$$

$$= \sum_{h: h_j \in \mathcal{H}_{y_j}} P(h|x;\theta) \quad (1)$$

where $P(h|x;\theta)$ is defined by a Gibbs distribution:

$$P(h|x;\theta) = \frac{\exp(\sum_k \theta_k \cdot F_k(h;\theta))}{Z(x;\theta)} \quad (2)$$

and $Z$ is the partition function which is defined by:

$$Z(x;\theta) = \sum_h \exp\left(\sum_k \theta_k \cdot F_k(h;\theta)\right)$$

Also $F_k$ is a summation over value of $f_k$ at each point of sequence and written as:

$$F_k(h,x) = \sum_{i=1}^{|x|} f_k(h_i, h_{i-1}, x, i) \quad (3)$$

Which $|x|$ is length of data sequence. Each feature function $f_k$ can be either a state function or pairwise function, the state function related to a single hidden variable in the model while pairwise function related to a pair of hidden variables.

### C. Connectionist Temporal Classification

Connectionist temporal classification (CTC) [14, 20], as its name suggests, is proposed for use in temporal classification tasks. It is used as output layer of RNNs for temporal classification problems.

Considering a training set $S$ (as explained in Section II.A), the objective function of CTC is defined as the sum of negative log-likelihood of samples $s_i = (x_i, y_i)$ in $S$:

$$J_{CTC} = -\sum_{s_i \in S} \log P(y_i|x_i) \quad (4)$$

For each frame of an input data sequence $x_i$ CTC receives the probabilities of all class labels from the unfolded RNN. Hence it receives a table of probabilities with size $t_i \times |\mathcal{Y}|$ as input (where $t_i$ is the length of the input sequence $x_i$) and efficiently computes $\log P(y_i|x_i)$ using dynamic programing. Correspondingly, it backproagates a table of errors to the bottom RNN.

III. PROPOSED METHOD

In the first section, we mentioned that one limitation of LDCRF model is that it can only be applied to sequence classification problems where training sequences are frame-wise labeled. CTC was introduced to train RNNs with unsegmented sequences.

In this section, we present a method for training LDCRF directly with unsegmented sequences. In our model, called CTC-LDCRF, we use CTC on top of LDCRF (see Figure 2). In the next subsections, we describe our model, its formulation and learning algorithm.

*A. Model Overview*

We propose a discriminative model for unsegmented sequence learning tasks. As shown in Figure 2, it consists of two layers: LDCRF layer and CTC layer. At the first step, LDCRF takes a data sequence as input and computes the label probabilities for each frame, then CTC takes both the LDCRF outputs and the desired label sequence and computes the gradient of its objective function with respect to label probabilities of the output nodes of the LDCRF model. Like the well-known backpropagation algorithm, we use the gradients returned by CTC to update the parameters of the LDCRF model. Finally, the gradient descent technique is used to optimize our model.

*B. CTC-LDCRF learning algorithm*

We assume that the training set consists of n pairs $(x_1, z_1), \ldots, (x_n, z_n)$ of data sequences $x_i = \{x_{i1}, \ldots, x_{it_i}\}$ along with their corresponding label sequences $z_i = \{z_{i1}, \ldots, z_{it_i}\}$ with unknown alighnment (which is explained in the Section II.A).

We use the following objective function to train our model with parameter set $\theta$:

$$J = -\sum_{i=1}^{n} \log P(z_i|x_i;\theta) \quad (5)$$

We use gradient descent for finding the optimal parameter values of the model. Let $\theta_k$ be the parameter that is associated with the state feature function $f_k$ in the model. By chain rule, for a training sample $i$ we have

$$\frac{\partial \log P(z_i|x_i)}{\partial \theta_k} = \sum_{j=1}^{t_i} \sum_{a \in \mathcal{Y}} \frac{\partial \log P(z_i|x_i)}{\partial P(y_{ij} = a|x_i)} \cdot \frac{\partial P(y_{ij} = a|x_i)}{\partial \theta_k} \quad (6)$$

As we described in Section II.C, the first term in the inner summation in (6) is computed by CTC, so we only need to calculate the second term which is the marginal conditional probability of (1):

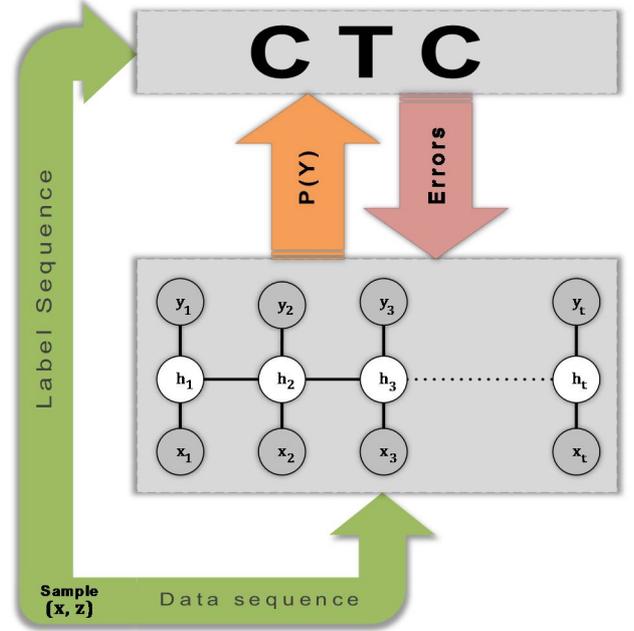

Figure 2. Illustration of our proposed method.

$$P(y_{ij} = a|x_i;\theta) = \sum_{s \in \mathcal{H}_a} P(h_j = s|x_i;\theta)$$

$$= \frac{\sum_{h:h_j \in \mathcal{H}_a} P(h|x_i;\theta)}{\sum_h P(h|x_i;\theta)} \quad (7)$$

The gradient of (7) with respect to $\theta_k$ can be written as (details omitted for space):

$$\frac{\partial P(y_{ij} = a|x_i)}{\partial \theta_k} = \sum_{b \in \mathcal{H}} \Big( P(h_j = b|x_i;\theta) \cdot f_k(j, x_i) \cdot \big(\delta_{b,k} - P(h_j = b|x_i)\big) \Big) \quad (8)$$

where $\delta_{b,k}$ is defined as:

$$\delta_{b,k} = \begin{cases} 1, & b = k \\ 0, & otherwise \end{cases}$$

combining these results, we have:

$$\frac{\partial J}{\partial \theta_k} = -\sum_{i=1}^{n} \sum_{j=1}^{t_i} \sum_{a \in \mathcal{H}} \Bigg( \frac{\partial J}{\partial P(h_j = a|x_i)} \cdot f_k(j, x_i) \cdot P(h_j = a|x_i) \cdot \big(\delta_{a,k} - P(h_j = a|x_i)\big) \Bigg) \quad (9)$$

The marginal probabilities $P(h_j = a|x)$ in (9), can be efficiently computed for all hidden states of the model using belief propagation. The gradient of $J$ with respect to the parameters of the pairwise potentials can be obtained in a similar manner.

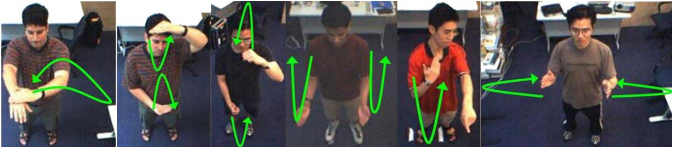

Figure 3. Illustration of the six arm gestures in the **ArmGesture** database [1]. Green arrows are indicative of the gesture trajectory for each class.

*C. CTC-LDCRF inference*

For testing, given a new test sequence $x$, we want to estimate the most probable label sequence $y^*$ using the formula:

$$y^* = \arg\max_y P(y|x, \theta^*) \qquad (10)$$

where $\theta^*$ is the set of parameter values that have been learned during the training phase.

## IV. EXPERIMENTs

In this section, we describe the data sets and the methodology which were used in our experiments to evaluate our model performance.

**Avatar-Eye Dataset:**

This dataset contains sequences of eye gestures gathered from six individual contributors interacting with an avatar. The goal is to discriminate eye gaze aversion gestures from other eye gestures for each frame of unsegmented video.

For each video, eye gaze estimated using two view-based appearance models explained in [21], and two-dimensional feature vector obtained from each frame of video. For each video sequence, frames were tagged either as "eye gaze aversion" or "other eye gestures".

**ArmGesture-Continuous Dataset:**

This dataset includes unsegmented sequences of arm gestures which are created based on the original ArmGesture dataset. The original dataset contains 724 samples of six types of arm gestures shown in Figure 3, and sequences were collected from 13 contributors with an average of 120 samples per class, where each sample is a sequence of 20-dimensional feature vectors.

Each sample of the Arm-Gesture Continuous is generated by serializing 3 to 5 randomly selected samples of six different gesture classes from the original dataset. This dataset contains 182 samples, with a mean length of 92 frames.

*A. Models*

We compare three configurations of our proposed method with LDCRF, CRF, and HMM on ArmGesture-Continues and Avatar-Eye data sets.

**Conditional Random Field:** we trained a single chain CRF with its standard objective function and regularized term, we varied window size from 0 to 2.

**Latent-dynamic Conditional Random Field:** we trained a single chain LDCRF model with the objective function described in [6] We are differing the number of hidden states per label from 2 to 6 and the window size from 0 to 2.

**Hidden Markov Model:** we used an HMM for each class. Each HMM was trained with segmented subsequences where the frames of each subsequence all fitted to the corresponding class. This training set contained the same number of frames like the one used for training the CRF and LDCRF models excluding frames is group into subsequences according to their tag. The final model is build by combining HMMs together.

**CTC-LDCRF:** we trained a single chain CTC-LDCRF model with three type of configuration. For the first configuration, we train model with unsegmented label sequence which never done before on these two data sets.

For the second configuration, the model trained with framewised tagged sequences similar to LDCRF and CRF training data.

For the third configuration, we use a two-step training procedure. At the first, we obtain pre-trained weights by training the model on segmented subsequences (similar to the HMM training set). At the next step, we used the pre-trained weights for initializing the model parameters and trained it on unsegmented data sequences. Within training, we differ the number of hidden states per label for the model from 2 to 6 and window size from 0 to 2.

For all of these three configurations, we are adding one more class label into the model to use as "non-gesture" or "blank" class label which is required in the CTC layer.

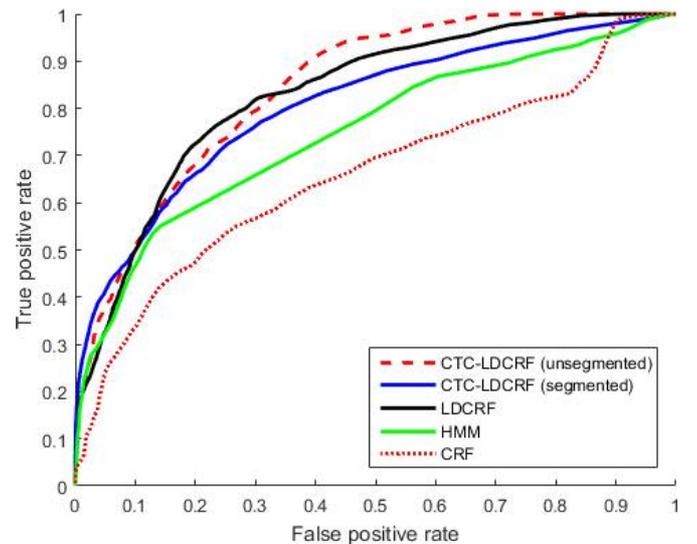

Figure 4. ROC curves from the Avatar-Eye experiment. Windows size set to one for all models.

TABLE I. EXPERIMETS RESULT FOR ARMGESTURE-CONTINUES DATASET

| Model | Configuration | Accuracy |
|---|---|---|
| CTC-LDCRF (w = 1, h = 3) | Pre training + Train with unsegmented sequence | 93.9% |
| CTC-LDCRF (w = 1, h = 3) | Train with unsegmented sequence | 89.5% |
| CTC-LDCRF (w = 1, h = 2) | Train with segmented sequence | 92.0% |
| LDCRF (w = 1, h = 3) | Train with segmented sequence | 92.2% |
| CRF (w = 1) | Train with segmented sequence | 89.1% |
| HMM | - | 91.2% |

### B. Methodology

For these two data sets, the experiments are carried out using the K-fold method for comparing models performance. In this method, the dataset is divided into K equal group and one of them chosen for the testing set while the other used for training and validation set. This procedure is repeated K times, and a mean of the K result reported for accuracy. We use K = 5 for the ArmGesture-Continues and K = 2 for the Avatar-Eye dataset. The measure for the model's performance is number of the correct predicted labels for each frame of test sequences:

$$accurency = \frac{\sum \#of\ correct\ labels}{\sum \#of\ sequence\ frames} \times 100 \quad (11)$$

### V. RESULTS AND DISCUSSION

We compare the result of the three configurations of our CTC-LDCRF learning algorithm on ArmGesture-Continues and Avatar-Eye data sets with standard LDCRF, Linear CRF, and HMM Models in our experiments. We use (11) to compare test results. The ROC curve for the test on Avatar-Eye dataset (Figure 4) shows that the CTC-LDCRF, which trains with unsegmented sequences, outperforms the other models. It can explain because of data sequences are frame-wise tagged manually by a human agent, and it may not be exactly correct especially on boundaries where label changed in sequence. Since we use unsegmented training sequence and don't use frame wised labels our method achieves better performance compared with other methods, also because of Avatar-Eye is a small dataset with 3-Dimentional feature vector and only 2 class label, the CTC-LDCRF method achieves good performance without needs to pre-trained.

Table I. shows the experimental results on ArmGesture-Continues dataset. The third configuration of Our CTC-LDCRF training methods has higher accuracy compare to others experimented methods.

### VI. CONCLUSION

In this paper, we develop a framework for training LDCRF model directly on the unsegmented sequence. We employed CTC approach in our framework to make the LDCRF model able to trains directly with unsegmented sequences. We did experiments on two gesture recognition data sets and showed that our model achieves better performance over other tested models, even though it is trained on unsegmented data sequence and gets less information than other models.

For the future works we plan to extend our framework and use a neural network such as LSTM and Convolutional Neural Network (CNN) as a feature extractor and make it possible to apply our framework on more complex, and raw real-world datasets [22-24].

ACKNOWLEDGMENT

We would like to thank Mostafa Rafiee, MSc student at the Ferdowsi University of Mashhad, for his noteworthy discussions and useful assist in implementations and experiments.